\pdfoutput=1

\documentclass[11pt]{article}

\usepackage[review]{acl}
\usepackage{lineno}
\usepackage{times}
\usepackage{latexsym}
\usepackage{graphicx}

\usepackage[T1]{fontenc}

\usepackage[utf8]{inputenc}

\usepackage{microtype}

\usepackage{inconsolata}

\usepackage{float}

\usepackage{natbib}

\usepackage{amsmath} 
\usepackage{booktabs}
\newcommand\headercell[1]{%
   \smash[b]{\begin{tabular}[t]{@{}c@{}} #1 \end{tabular}}}

\title{\huge Crafting Narrative Closures: Zero-Shot Learning with SSM Mamba for Short Story Ending Generation}

\author{
  Divyam Sharma \\
  University of Michigan \\
  {\tt divyams@umich.edu}
  \And
  Divya Santhanam \\
  University of Michigan \\
  {\tt divyasan@umich.edu}
}

\begin{document}
\maketitle
\nolinenumbers

\begin{abstract}
Writing stories is an engaging yet challenging endeavor. Often, authors encounter moments of creative block, where the path forward in their narrative becomes obscured. This paper is designed to address such moments by providing an innovative solution: A tool that completes stories based on given prompts. By inputting a short story prompt, users can receive a conclusion to their story, articulated in one sentence or more, thereby enhancing the storytelling process with AI-driven creativity. This tool aims not only to assist authors in navigating writer's block but also to offer a fun and interactive way for anyone to expand on story ideas spontaneously. Through this paper, we explore the intersection of artificial intelligence and creative writing, pushing the boundaries of how stories can be crafted and concluded. To create our final text-generation models, we used a pre-trained GPT-3.5 model and a newly created finetuned SSM-Mamba model, both of which perform well on a comprehensive list of metrics including BERT score, METEOR, BLEU, ROUGE, and Perplexity. The SSM model has also been made public for the NLP community on HuggingFace models as an open source contribution, which for the timebeing is a first of its kind state-space model for story-generation task on HuggingFace.
\end{abstract}

\section{Introduction}
Our paper aims to develop a sophisticated text-generator that effectively writes an applicable and readable conclusion to a short story prompt. While conducting research into similar papers, we found many instances of selection based models that chose the most relevant ending to a story from presented options. Our paper differs in that we sought out to generate the endings ourselves. Our text-generator requires four sentences of text input that represented the bulk of the story. From this text, our generator produces a one-sentence-long conclusion. The generator was trained on thousands of five-sentence-long stories that resembled the type of user input that we would expect to see. To achieve this, we used a total of four models, including a baseline Random Selection model, a baseline N-Gram model, a self fine-tuned SSM-Mamba model, and GPT-3.5. The latter two models produced the best results.

Our paper aims to solve the problem of writer's block. There are several instances where having a tool that can conclude a story would be beneficial. For example, a potential user of our text-generator could be an author looking for a conclusion to their short story, or a parent that is looking for a personalized bedtime story to tell their child. 

While doing this paper, we have learned about the complexities of prompt design while working with GPT-3.5, and experienced a significant learning curve while making use of efficient techniques for fine-tuning using LoRa, PEFT, and SFTTrainer. We also took a deep dive into state-space models, their working mathematics, and how they are changing the paradigm for text generation, currently influenced by transformers and attention. The one take away for everyone from our learning curve will be the utilization of Low Rank Adaptation and Parameter Efficient Fine Tuning which we have described in depth in further sections.

Our paper innovatively advances text generation by developing a tool that autonomously generates the concluding sentence of a short story, trained on diverse narrative structures. This not only assists in overcoming writer's block but also introduces a novel application of state-of-the-art AI techniques, including fine-tuning with LoRa, PEFT, and SFTTrainer, and explores the potential of state-space models in transforming text generation. We've made this sophisticated tool available open source, broadening access to cutting-edge AI benefits for creative writing and personal storytelling.

\section{Data}
The data we have used for our paper is five-sentence-long short stories. To obtain the short stories, we have accessed a ROCStories Corpora database maintained at Rochester University.

The dataset we used is from the Story Cloze Test and ROCStories Corpora which has 98,161 short five-sentence-long stories that can be found \href{https://cs.rochester.edu/nlp/rocstories/}{here}. The dataset has free access to anyone upon request, which we have already received. This paper aims to evaluate story prediction. Given two possible conclusions to a story, the model chooses the more favorable option. Our paper extends this concept by focusing on text generation over prediction. Rather than being given options to choose from, our model generates the text itself.

The ROCStories Corpora consists of data obtained during the spring of 2016 and the winter of 2017. The quantity of observations from both datasets can be seen in Table 1. 

\begin{table}[H]
    \centering
    \begin{tabular}{|| c c ||}
    \hline
    ROCStories 2016  & ROCStories 2017 \\ [0.5ex] 
    \hline\hline
    45496 & 52665 \\
    \hline
    \end{tabular}
  \caption{Quantity of observations in ROCStories Corpora}
\end{table}
\vspace{-2em}

For the purposes of this paper, we have treated the two datasets as one. The datasets consist of five sentences for each observation. In building our final model, after training, we have used four sentences as input and evaluated the output against the actual fifth sentence in the story.

An example observation from our dataset is shown in Table 2.

\begin{table}[H]
    \begin{tabular}{| p{0.35\linewidth} | p{0.6\linewidth} |}
    \hline
    Category & Data \\ [0.5ex] 
    \hline\hline
    ID & 9a51198e-96f1-42c3-b09d-a3e1e067d803 \\
    \hline
    Title & Overweight Kid  \\
    \hline
    Sentence1 &  Dan's parents were overweight. \\
    \hline
    Sentence2 &  Dan was overweight as well. \\
    \hline
    Sentence3 &  The doctors told his parents it was unhealthy. \\
    \hline
    Sentence4 & His parents understood and decided to make a change.  \\
    \hline
    Sentence5 &  They got themselves and Dan on a diet. \\
    \hline
    \end{tabular}
  \caption{An example observation from the ROCStories dataset.}
\end{table}

\section{Related Work}

A similar work that has been done by \cite{mostafazadeh2016corpus} at the University of Rochester has focused on coming up with a benchmark - Story Cloze benchmark as a method for evaluating the understanding of short (five-sentence) commonsense stories. In the
Story Cloze benchmark, the first four sentences of an everyday story are given, and the concluding sentence is withheld. The model is offered two choices, both of which are contextual to the preceding sentences but one of which is right and the other is wrong. So the focus of the above-mentioned work is on predicting the right ending out of the two options.
Other works on similar lines focusing on predicting the right ending to solve ROCStory Cloze Task have been such as \cite{schwartz2017story}. To improve the performance, features like topic words and sentiment score are also extracted and incorporated \cite{chaturvedi2017story}. Neural network models have also been applied to this task (e.g., \cite{huang2013learning} and \cite{cai2017pay}), which use LSTM to encode different parts of the story and calculate their similarities. In addition, \cite{li2018multi} introduces event frame to their model and leverages five different embeddings. Finally, \cite{radford2018improving} develops a transformer model and achieves state-of-the-art performance on ROCStories, where the transformer was pre-trained on BooksCorpus (a large unlabeled corpus) and finetuned on ROCStories.

Our intended work focuses on being able to generate the whole last line of the 5-sentence story, aiming to have contextual meaning and logical coherence with the rest of the story (the 4 seen sentences) and not just selecting the right ending choice out of a given few options.

\section{Methods}

\subsection{Preprocessing Steps:}
\begin{itemize}
    \item \textbf{Tokenization:} Split the text into tokens (words or subwords) using a tokenizer suited for the chosen model architecture (implemented using the AutoTokenizer Class with model = \{ GPT-3.5, state-space/Mamba-370m-hf \}).
    \item \textbf{Cleaning:} Removed and corrected typographical errors, standardized quotation marks, and handled special characters to ensure text consistency.
    \item \textbf{Segmentation:} Divided each story into two parts: the body (beginning and middle) and the ending using [SEP] token. This segmentation facilitated the model to learn the transition from the story body to its ending.
    \item \textbf{Vectorization:} Converted tokens into numerical vectors using the tokenizer's vocabulary. This step is crucial for feeding textual data into neural networks.
\end{itemize}

\subsection{Model Selection and Architecture}
\subsubsection{ Model 1: Large Language Model - GPT-3.5}
\begin{itemize}
    \item \textbf{Pre-trained Language Model:} We loaded the transformer-based pre-trained language model GPT-3.5, using credits that have been offered by the University, because of the model's proven capacity for generating coherent text. The model choice was a compromise between limited computational resources and desired output quality. The openai library\footnote{\href{https://github.com/openai/openai-python}{openai Library}} was used to load the model.
    \item \textbf{Prompt Design:} We used rigorous testing to determine the best performing prompt. We utilized Chain-Of-Thought prompting and In-Context Learning to compare performance. Chain-Of-Thought prompting had the model explain its reasoning after generating a conclusion for a given story to improve performance. In-Context Learning prompting gave the model multiple examples to base its solution off of. We used human evaluation to evaluate initial performance of the different prompts. The final prompts that were tested had seemingly equal quality of performance. Considering the trade offs between computational resources and quality of results, the prompt that was chosen was the most succinct and straightforward in order to best preserve resources. Specifically, the prompt script that was chosen followed the following format: 'Write a conclusion to the following story:', concatenated with the first four sentences of the given story.
\end{itemize}

\subsubsection{Model 2: Selective State-Space Model (SSM) - Mamba}

\begin{itemize}
    \item \textbf{Pre-trained SSM Mamba Model:} Fine-tuned Mamba \cite{gu2023mamba} on our dataset of short stories, focusing on teaching the model to understand how narratives evolve and conclude. This step customizes Mamba's general capabilities to our specific task of generating story endings.
    \item \textbf{Fine-tuning Parameters:} Used a learning rate of 2e-3, with a batch size of 8, for 1 epoch. Employed a linear scheduler for learning rate decay.
    \item \textbf{Special Tokens:} Incorporated special tokens [SEP] or markers to delineate different sections of the stories (beginning + middle, end) and significant narrative shifts. This aided the model in understanding the structure of narratives.
\end{itemize}

\begin{figure}[H]
    \centering
    \includegraphics[width=\linewidth]{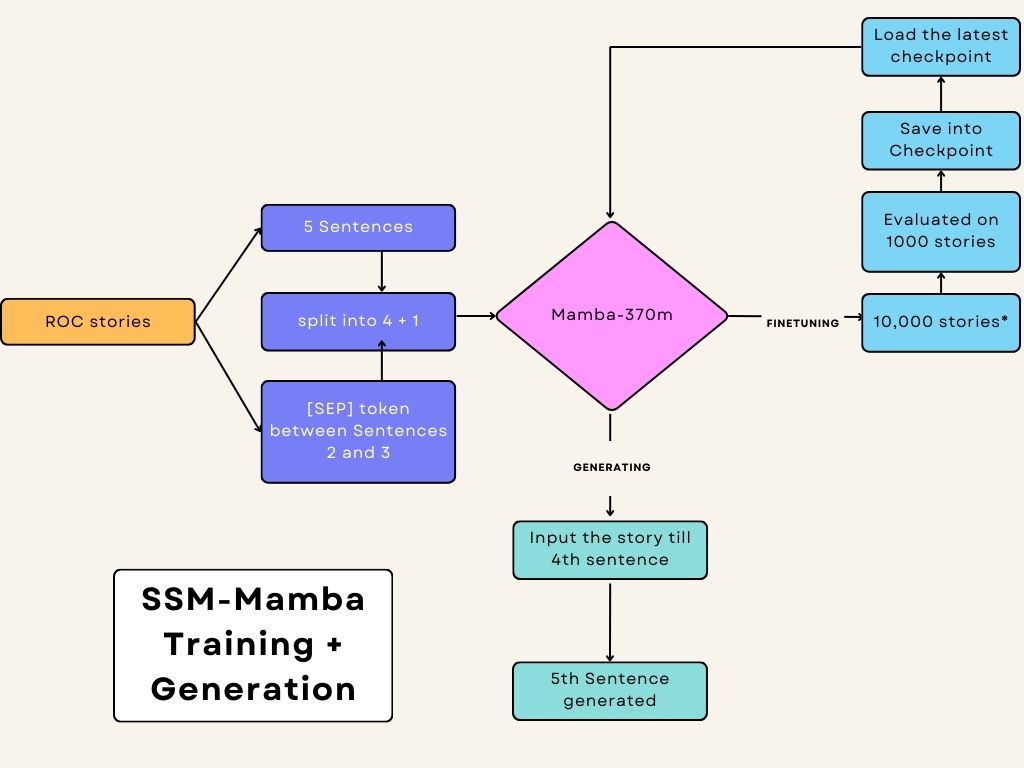}
    \caption{SSM-Mamba Training and Generation}
    \label{fig:Model Training Flow}
\end{figure}

\textbf{Rationale for SSM:} Opting for a state-space model like Mamba leverages recent advancements in NLP to capture the temporal and dynamic aspects of story progression, which is crucial for generating coherent and contextually relevant story endings.
\begin{itemize}
    \item \textbf{State Representation:} Define the states of our narrative. In the context of SSMs, a state could represent various elements of the story (e.g., plot points, character development stages) encoded in a vector form.
    \item \textbf{Dynamics Model:} This component models how the story progresses from one state to another. Using the [SEP] token, it would transition from narratives in the beginning towards the ending.
    \item \textbf{Observation Model:} Defines how the observed outputs (e.g., sentences or paragraphs of the story) are generated from the hidden states. This model helps in generating the text of the story ending based on the final states reached by the dynamics model.
\end{itemize}

\subsubsection{Training Procedure}
\textbf{Finetuning Mamba-370m} \\
\textbf{Why Mamba-370m:} \\ Attempts were made to finetune Mamba-2.8B, Mamba-1.4B and Mamba-790M but due to memory constraints of having to use only one GPU on University GPU Cluster, the maximum number of parameters that could be loaded were 370M only better than the Mamba-140M.\\ \\ \textbf{PEFT Training}:\\ PEFT \footnote{\href{https://huggingface.co/docs/peft/en/index}{PEFT}} (Parameter-Efficient Fine-Tuning) serves as a library that allows for the effective adaptation of large pre-trained models to specific applications without the need to fine-tune every parameter. This approach is crucial as full parameter fine-tuning can be excessively expensive. By adjusting only a select set of additional parameters, PEFT 2 reduces both computational and storage demands significantly. Consequently, this facilitates comparable performance to models that are fully fine-tuned, enhancing the feasibility of training and maintaining large language models (LLMs) on consumer-grade hardware.
    \begin{enumerate}
        \item \textbf{Error Correction Focus:} PEFT emphasizes correcting previous errors during the training process, which can lead to more robust models.
        \item \textbf{Incremental Learning:} By progressively adjusting the model, PEFT supports gradual learning, making it easier to manage and fine-tune.
        \item \textbf{Reduced Overfitting:} The methodology inherently counters overfitting by continuously integrating error feedback, which helps improve generalization.
        \item \textbf{Efficiency in Training:} PEFT can be more efficient than standard backpropagation as it focuses specifically on error-prone areas, potentially reducing training time.
    \end{enumerate}
    
\textbf{LoRa - Low Rank Adaptation: Enhancing Parameter Efficiency} 

The LoRa \footnote{\href{https://huggingface.co/docs/diffusers/main/en/training/lora}{LoRa}} configuration within Parameter Efficient Finetuning (PEFT) involves incorporating Low-Rank Adaptation (LoRa) techniques into the PEFT framework. LoRa, by introducing low-rank matrices, modifies only a subset of a model's parameters, thereby making the adaptation process more parameter-efficient and targeted. Here’s a deeper look into the integration of LoRa within PEFT and the advantages it offers:
    \begin{itemize}
    \item \textbf{Minimal Parameter Updates:} LoRa fine-tunes large models by updating only additional low-rank matrices, reducing the number of trainable parameters significantly.
    \item \textbf{Resource Efficiency:} This efficiency is crucial for both PEFT and SFTTrainer as it minimizes computational and memory requirements, enabling more sustainable training processes.
    \end{itemize}

\textbf{SFTTrainer}
\begin{enumerate}
    \item \textbf{Task-Specific Fine-Tuning:} SFTTrainer \footnote{\href{https://huggingface.co/docs/trl/en/sft_trainer}{SFTTrainer}}  is designed to fine-tune models on specific tasks, leading to improved performance on targeted applications.
    \item \textbf{Flexibility:} It offers the flexibility to adjust models according to the unique characteristics and requirements of different data sets or tasks.
    \item \textbf{Resource-Efficient:} Optimized for lower resource consumption compared to training a model from scratch, making it suitable for environments with limited computational power such as our Training Use case on the in-house GPU cluster.
\end{enumerate}
\textbf{Environment:} Trained the model using a GPU-accelerated environment (in-house GPU cluster) to handle the computational load efficiently. As shown in Figure 1, due to timeout constraints on in-house GPU cluster, the model was trained for 10,000 stories in one go, saved as checkpoint and picked up from there for further training using a SLURM dependency usage. \\
\textbf{Loss Function:} Used Cross-Entropy Loss to calculate the difference between the generated endings and the actual endings, optimizing the model's ability to predict the next token accurately.\\
\textbf{Optimization:} Utilized the AdamW optimizer for adjusting model weights, reducing the loss over training epochs.

\section{Evaluation and Results}
Evaluating story-ending generation requires a nuanced approach to gauge both the technical and creative quality of generated narratives. Therefore, we introduce here various evaluation methods which we used to test our models on, encompassing automated metrics like BLEU, ROUGE, METEOR, BERTScore, and Perplexity, which quantitatively measure aspects such as n-gram overlap, semantic similarity, and fluency. Additionally, human evaluation will be important for looking into coherence, creativity, and emotional impact that automated metrics might overlook. Together, these evaluation strategies provided us with a comprehensive assessment framework, ensuring a balanced analysis of a model's ability to generate compelling and contextually fitting story endings.

As an initial baseline approach, we used a random selection model. First, we created a fifth-sentence database obtained by pulling directly from each short story in the original dataset. Then, for each observation in the data, we randomly selected a concluding sentence from the fifth-sentence database and evaluated the predictions against the ground truth sentences using BERT, METEOR, BLEU, ROGUE, and Perplexity. The average metric results for all 98,161 observations are listed in the Table 3. 

As a second baseline, we used a variety of multigram models, specifically with orders 2, 4, 7, and 10. Our final evaluation was performed on the 10-gram model. The multigram models used a limited number of contextual characters to predict the following character. Using a specified contextual window, the models trained on all of the existing stories in the database and predicted 100 sentences that ideally followed the grammatical structure and subject matter of the training data. The predicted sentences were compared against 100 real sentences from the database. The results of the 10-gram model's performance are listed in Table 3.

\begin{table}[H]
\noindent
    \begin{tabular}{@{} *{3}{c} @{}}
    \hline
    \headercell{Evaluation \\ Metric} & \multicolumn{2}{c@{}}{Baseline} \\
    \cmidrule(l){2-3}
    & Random Selection & N-gram (n=10) \\ \hline
    BERT & 0.869 & 0.852 \\ \hline
    METEOR & 0.083 & 0.053 \\ \hline
    BLEU & 0.001 & 0.000 \\ \hline
    ROUGE & 0.070 & 0.053 \\ \hline
    Perplexity & 140.054 & 518.247 \\ \hline
    Human Score (1-5) & 1 & 1.5 \\ \hline
    \end{tabular}
\caption{Average metric scores for the random selection and multigram baseline models. \\ *BERT score listed is F1 score and ROUGE score listed is ROUGE1.}
\end{table}

We used two final models for our story-generation task. Our first model was a fine-tuned SSM-Mamba model trained on approximately 40,000 data points from the short story corpus. The model was tested on approximately 2,000 data points. The results for this model are shown in Table 4.

\begin{table}[H]
\noindent
    \begin{tabular}{@{} *{3}{c} @{}}
    \hline
    \headercell{Evaluation \\ Metric} & \multicolumn{2}{c@{}}{Final Models} \\
    \cmidrule(l){2-3}
    & SSM-Mamba & GPT-3.5 \\ \hline
    BERT & 0.878 & 0.878 \\ \hline
    METEOR & 0.100 & 0.206 \\ \hline
    BLEU & 0.013 & 0.013 \\ \hline
    ROUGE & 0.180 & 0.186 \\ \hline
    Perplexity & 171.000 & 48.881 \\ \hline
    Human Score (1-5) & 3.9 & 4.2 \\ \hline
    \end{tabular}
\caption{Average metric scores for the SSM-Mamba and GPT-3.5 models. \\ *BERT score listed is F1 score and ROUGE score listed is ROUGE1.}
\end{table}

The hyperparameters that led to a decent performance of the SSM-Mamba model have been a higher learning rate of 2e-3 and LoRa configuration in PEFT training while using the SFTTrainer as the wrapper trainer. With a slower learning  rate, since we were limited by GPU time, the training was not as fast and hence we would have to do a significantly more amount of training in time which was handled well with a higher training rate accompanied with L2 Regularization factor of 0.01. This combined with an efficient training as described in methodology using LoRa configuration on PEFT and SFTTrainer has led to satisfying results.

The second model we used was GPT-3.5. Through rigorous prompt design including Chain-Of-Thought and In-Context Learning prompting, we used human evaluation to determine the best performing prompt, and proceeded to generate story conclusions for approximately 3,000 data points. The results for this model are shown in Table 4.

Outlined below are the evaluation metrics and their significance in the context of our baseline models as well as final implementation.

\subsection{BLEU Score \cite{papineni2002bleu}} 
To apply the BLEU score for evaluating a story-ending generation task where we have the actual 5th sentence of the story and a generated 5th sentence, we'll compare the generated sentence against the actual sentence using the BLEU metric. BLEU will quantify how close our generated ending is to the actual ending based on n-gram overlap.

\subsection{ROUGE \cite{lin2004rouge}}
The ROUGE Score and its variants (Rouge-N/L/W) facilitate evaluation by measuring the overlap of n-grams, the longest common subsequences, and skip-bigrams between the generated text and a set of reference texts. For story ending generation, ROUGE can assess the extent to which key phrases and narrative elements in the generated story ending align with those in the actual story ending, offering a quantitative measure of narrative fidelity.

\subsection{METEOR \cite{banerjee2005meteor}}
In the context of story ending generation, METEOR provides a nuanced assessment of how well the generated ending captures the meaning and fluency of the actual story ending, taking into account paraphrasing and flexible expression.

\subsection{BERTScore \cite{zhang2019bertscore}}
BERTScore leverages the contextual embeddings from models like BERT to compute the semantic similarity between the generated text and the reference text. By comparing the cosine similarity of embeddings for matched words, BERTScore offers an evaluation of semantic congruence between the generated story ending and the actual ending, highlighting the model's ability to generate semantically relevant and contextually appropriate narrative conclusions.

\subsection{Perplexity \cite{jelinek1977perplexity}}
Perplexity measures the uncertainty of a language model in predicting the next token, providing an indication of the fluency and naturalness of the generated text. For story ending generation, a lower perplexity score on the generated ending suggests that the text is more predictable and fluent, reflecting the model's linguistic competency in crafting coherent and contextually fitting narrative closures.

\subsection{Human Evaluation}
Despite the advancement in automated metrics, human evaluation remains indispensable for assessing creative text generation tasks such as the story ending generation. Human judges - the authors of this work each evaluated 225 one sentence generated endings for coherence, narrative satisfaction, creativity, emotional impact, and grammatical correctness, looking deeply into the qualitative aspects of story generation that automated metrics cannot fully capture.

\section{Discussion}

As seen in Table 3, both baseline models had similar scores for each evaluation metric. On average, the random selection model appeared to marginally outperform the multigram model in each metric category. Notably, both models produced fairly high BERT scores and fairly low METEOR, BLEU, and ROUGE scores. This indicates that the generated text itself is semantically similar to the rest of the stories, but not too similar contextually or with respect to ordering. The perplexity of the multigram model was much higher than that of the random selection model, indicating that the readability of the random selection model output was much better. However, since the sentences were randomly selected, they didn't make a logical ending to the story and hence a lower human score as the coherence didn't hold. This would make sense, considering the random selection model was choosing from a pool of preexisting fifth-sentences from the original dataset, while the multigram model was generating the text character by character. Overall, the random selection model was the better performing baseline.

As seen in Table 4, both the finetuned SSM-Mamba and GPT-3.5 models significantly improved on the baseline results. Both models had similar BERT scores, METEOR scores, BLEU scores, and ROUGE scores, with GPT-3.5 marginally outperforming SSM-Mamba. The perplexity score of SSM-Mamba was lower than that of both GPT-3.5 and the random selection baseline, but still beat the N-gram baseline. The best overall perplexity score was achieved by GPT-3.5, indicating a higher readability, fluency, and contextual similarity. Table 5 shows an example story from the corpus, the target concluding sentence, and the endings generated by both SSM-Mamba and GPT-3.5.

\begin{table}[H]
    \begin{tabular}{| p{0.35\linewidth} | p{0.6\linewidth} |}
    \hline
    Category & Data \\ [0.5ex] 
    \hline\hline
    Sentence1 &  Nick's old smart phone was very slow. \\
    \hline
    Sentence2 &  He researched his options for a new smartphone. \\
    \hline
    Sentence3 &  Nick went to the store. \\
    \hline
    Sentence4 & He purchased a much faster smartphone.  \\
    \hline
    Target &  Nick is much happier with his new device. \\
    \hline
    Finetuned SSM-Mamba &  Nick was very happy with the new phone. \\
    \hline
    GPT-3.5 &  As soon as Nick turned on his new smartphone, he immediately noticed the vast improvement in speed and efficiency. \\
    \hline
    \end{tabular}
  \caption{An example story from the ROCStories dataset and the generated endings for each final model.}
\end{table}

As seen in Table 5, both the SSM-Mamba model and GPT-3.5 produced very relevant conclusions, with similar sentiment to the target conclusion.

While just starting, the SSM-Mamba model without any finetuning was unable to produce any endings and was just repeating the context of the story given as the output. Then iteratively as the model was finetuned on the data, it improved. As highlighted in the Methods in detail, the LoRa config in PEFT training with SFTTrainer as the wrapper trainer worked wonderfully for the limited resources for finetuning. The State-Space Model, famous and efficient for its selective learning, was able to learn the nuances of shift between the starting line of the stories to the middle part of the short stories and then shifting to the concluding line. As a result, the finetuned SSM-Mamba performs well on all metrics, remarkably so for the limited training that it got, except perplexity which we infer is a result of not learning the writing style in depth due to lesser training.

However, in the training effort, we were limited by the in-house GPU cluster timeout and 1 GPU limit per request. So we had to come up with a smart SLURM dependency usage and flow of training while saving checkpoints and reloading the last saved checkpoint for further training. This pushed us to make use of more refined Training configurations. However, as a sneak-peek into future, attempt should be made to finetune the bigger Mamba models on the dataset, which we believe could probably outperform the GPT-3.5.

\section{Conclusion}
When we began this paper, we sought out to create a text-generation model that would write a conclusion to a given short story. The purpose of creating such a model was to try and help solve the problem of writer's block. Over the course of the paper, we created a Random Selection baseline model, an N-Gram baseline model, fine tuned the SSM-Mamba model, and used GPT-3.5 with efficient prompting. With each new model we used, we saw an improvement in performance. While GPT-3.5 performed the best, SSM-Mamba was a competitive alternative. Looking forward, we would like to see if training GPT-3.5 on data from the short story corpus and fine tuning would result in a significant improvement in performance. \\ \\
\textbf{Model Deployment on HuggingFace}\\
 HuggingFace models has a lot of models fine-tuned for story-writing task based on the LLM GPT-2 architecture but there was none until now which was fine-tuned on Mamba SSM. Since we have come up with a reasonable performing fine-tuned Mamba-based SSM model for story generation, we have put it up for use on HuggingFace for the active NLP community and have proudly contributed to Open Source as well, as a culmination of the research. This is the first and only state-space model on HuggingFace for the time-being which is finetuned for story generation task. The Mamba-370m-story-generation model is \href{https://huggingface.co/DdIiVvYyAaMm/mamba-370m-story-generation}{available publicly here on HuggingFace}.

\section{Other Things We Tried}
Though we were able to produce desirable results, we faced several roadblocks along the way. For example, while attempting to fine tune the SSM-Mamba model, we were unable to fine tune on the 2.8B parameter model like we had originally planned. This was due to insufficient memory and limited access to the in-house GPU cluster GPUs. We attempted fine tuning on the 1.4B parameter model and the 790M parameter model. Ultimately, the largest number of parameters we were able to load was 370M. Also, coming up with the perfect concise short prompt for finetuning for SSM-Mamba took some experiments. The final short prompt has been to just follow the story sentences with "Complete this story by generating its last line to give it a logical ending:". Figuring out the exact LoRa configurations was an exciting adventure of experiments as well to determine the right rank and target modules.
Another roadblock we faced was with the GPT-3.5 model. Ideally, we would have liked to train the model on our short story corpus, but we realized pretty quickly that this would be impossible given the limited funds we had access to. As a result, we used the pre-trained model. As expected, the model still performed very well. 

\section{What We Would Have Done Differently or Next}
While both of our final models improved significantly on the baselines, there are several ideas we were not able to implement that we believe would've resulted in even better performance. 

The biggest obstacle we faced in developing our models was limited resources. While we were able to significantly increase speed through the use of the GPUs provided by the University, we lacked the funds necessary to train/test our data to the lengths we would've liked. If we had more resources, we would have fine tuned the GPT-3.5 model on our short story corpus to produce even better results. Due to a low amount of funds, we were limited to using the pre-trained model. Additionally, we were only able to test the GPT-3.5 model on approximately 3,000 data points. Ideally, we could have utilized all 98,000+ stories in our database.

Another obstacle, especially for Mamba training we faced was a lack of GPU time. Due to the limited time allowance on the use of the campus GPUs (limited to 1 GPU per job), training the SSM-Mamba model took much longer than anticipated. During the training process, we had to wait for our submitted jobs to run out of time, save the current checkpoint, load the last saved model checkpoint, and submit a new job building off of said checkpoint. Ideally, we would be able to train the entire model in one job.


\bibliography{custom}

\end{document}